\documentclass[conference]{IEEEtran}
\IEEEoverridecommandlockouts

\usepackage{cite}
\usepackage{amsmath,amssymb,amsfonts}
\usepackage{algorithmic}
\usepackage{graphicx}
\usepackage{textcomp}
\usepackage{xcolor}
\usepackage{hyperref}
\usepackage{multicol,multirow}
\usepackage{makecell}
\usepackage{tikz}
\usetikzlibrary{shapes,arrows,positioning,calc}

\def\BibTeX{{\rm B\kern-.05em{\sc i\kern-.025em b}\kern-.08em
    T\kern-.1667em\lower.7ex\hbox{E}\kern-.125emX}}

\begin{document}

\title{Multi-Adapter PPO: A Cross-Attention Enhanced Wavelength Selection Framework for LIBS Quantitative Analysis}
\author{\IEEEauthorblockN{1\textsuperscript{st} Hao Li}
	\IEEEauthorblockA{\textit{Electrical and Computer Engineering} \\
		\textit{University of Arizona}\\
		Tucson, USA \\
		\texttt{lihao@arizona.edu}}
	\and
\IEEEauthorblockN{2\textsuperscript{nd} Man Fung Zhuo}
\IEEEauthorblockA{\textit{Electrical and Computer Engineering} \\
	\textit{University of Arizona}\\
	Tucson, USA \\
	\texttt{zhuomanf@arizona.edu}}
}
\maketitle
\begin{abstract}
	Laser-induced breakdown spectroscopy (LIBS) quantitative analysis faces critical challenges in wavelength selection due to high-dimensional spectral data and the fundamental trade-off between prediction accuracy and feature efficiency. This paper presents a novel Multi-Adapter PPO framework that transforms wavelength selection into a reinforcement learning problem, leveraging cross-attention mechanisms and multiple specialized adapters to capture complex spectral relationships. Our approach outperforms traditional Particle Swarm Optimization (PSO) by an average of 28.4\% in comprehensive score and 45.2\% in prediction accuracy across steel and coal datasets. The proposed method demonstrates superior performance in balancing prediction accuracy with feature efficiency, achieving state-of-the-art results in LIBS quantitative analysis while maintaining interpretability and computational efficiency. We released our code and dataset here: https://github.com/Hflying/MAPPO
\end{abstract}
\begin{IEEEkeywords}
Laser-induced breakdown spectroscopy, wavelength selection, reinforcement learning, Multi-Adapter network
\end{IEEEkeywords}
\section{Introduction}
\label{sec:intro}
Laser-induced breakdown spectroscopy (LIBS) has established itself as a versatile analytical technique for elemental detection and quantification, with applications spanning environmental monitoring, materials science, and industrial quality control. 
High-accuracy quantitative measurement via LIBS mainly deal with two critical challenges, spectral noise\cite{yan2025review,jiang2021baseline,shen2022single,gilda2019automatic,kazemzadeh2022cascaded,jiao2024three,he2023new,shang2024study} and data scarcity\cite{qiao2018protein,harefa2021performing,wei2020variations}. 
In LIBS, each element emits characteristic spectral peaks at specific wavelengths, which serve as the cornerstone for both qualitative identification and quantitative measurement. However, extracting reliable relationships between these spectral features and true elemental concentrations is complicated by matrix effects, overlapping peaks, and non-linear correlations—factors that traditional data processing pipelines often struggle to address comprehensively. Dimensionality reduction methods, such as principal component analysis (PCA)\cite{vrabel2020restricted,yuan2021rapid,ma2023step,liu2025learning}, while effective in simplifying data complexity, often obscure the direct association between individual characteristic peaks and their corresponding elemental concentrations by transforming the original spectral space.
The physics-driven interpretation of spectral features\cite{ding2018hybrid,wang2024application} and their relationship to elemental composition provides not only theoretical validation but also insights for improving the robustness and interpretability of quantitative models. 
Conversely, conventional wavelength selection algorithms, despite their focus on retaining informative wavelengths, frequently underperform in LIBS due to their inability to adapt to the dynamic and complex nature of spectral data, as highlighted by limitations in rigid thresholding or linear feature weighting approaches.

Therefore, a critical challenge in LIBS quantitative analysis lies in wavelength selection, as raw spectral data is inherently high-dimensional, containing redundant information, background noise, and spectral interferences that hinder accurate concentration prediction. 
Variable selection methods have been proposed including Mutual Information\cite{vergara2014review}, Chi-squared test, and Information Gain (IG), (they evaluate feature relevance based on statistical measures and are computationally efficient) and Particle Swarm Optimization \cite{yan2019novel}, Randomization Test\cite{xu2009wavelength}, and Genetic Algorithm\cite{pontes2009classification} (they consider the relationship between feature subsets and learning models).
Particle swarm optimization (PSO)\cite{he2021quantitative,wang2024application}, a widely adopted swarm intelligence algorithm, has been widely employed for wavelength selection in LIBS, leveraging collective search dynamics to identify optimal feature subsets. However, traditional PSO and its hybride variants\cite{yun2019overview,saptoro2012modified,li2026golden,li2021latency,li2026revisiting,li2026r,xu2017reliable} suffer from limitations such as premature convergence and suboptimal precision when navigating the high-dimensional and noisy landscape of LIBS spectra, where subtle but critical peaks (e.g., those from trace elements) are easily overshadowed by stronger signals or noise. 


\begin{figure}[htbp]
	\centering
	\includegraphics[width=1.0\linewidth]{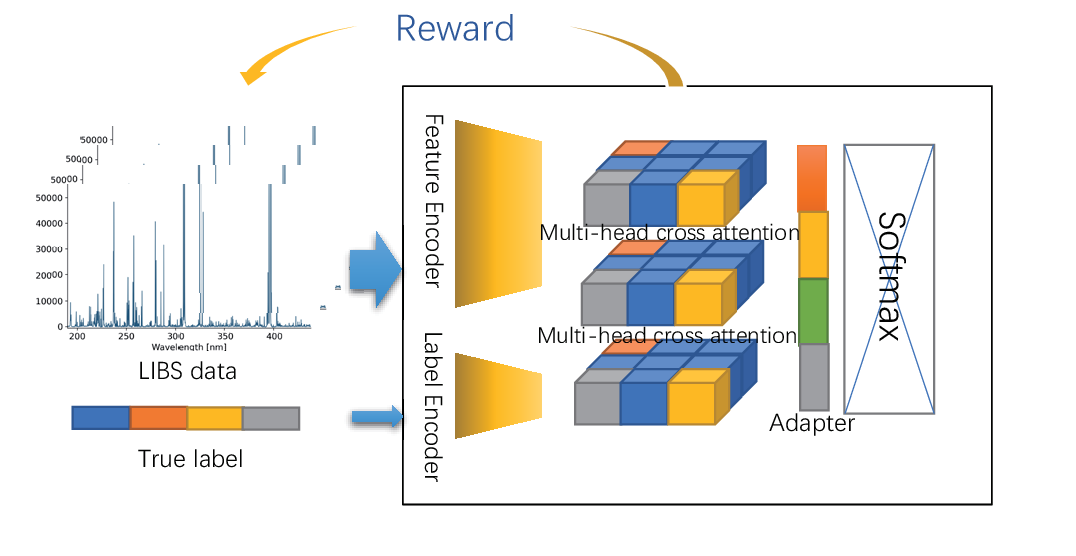}
	\caption{Overview of Multi-Adapter PPO Architecture. The framework consists of dual encoders (feature and target) that process spectral data and target variables, followed by multi-head cross-attention to capture spectral–target relationships. Four specialized adapters learn diverse feature-target mapping patterns, which are aggregated through learnable weights. The final policy network outputs action probabilities for wavelength selection or stopping.}
	\label{fig:architecture}
\end{figure}

If we delve into the essence of wavelength selection in LIBS spectral analysis, it inherently involves sequential decision-making—where each choice of wavelength affects subsequent selections—and requires optimizing a specific objective (e.g., enhancing signal informativeness while reducing noise), which aligns perfectly with the core paradigm of reinforcement learning, where an agent learns to make sequential decisions to maximize cumulative rewards.
A similar insight has been applied in the hyperspectral band selection domain. For instance, in \cite{feng2021deep}, the authors transform the hyperspectral band selection task into a reinforcement learning problem, proposing an A2C-based algorithm and leveraging a semi-supervised EvaluateNet to assess the efficiency of selected bands. This work validates the feasibility of framing spectral selection tasks within a reinforcement learning framework, laying a foundation for further algorithmic optimizations.
However, it is widely recognized that the A2C algorithm has limitations in terms of generality compared to more advanced alternatives. In contrast, Proximal Policy Optimization (PPO) offers distinct advantages in handling sequential decision-making and complex optimization tasks. By integrating an actor-critic framework with efficient, clipped policy updates, PPO can stably and adaptively prioritize informative wavelengths in LIBS spectral analysis while suppressing noise and interference. This not only addresses the core demands of wavelength selection but also overcomes the generality constraints of A2C, making it far better suited for the nuanced and variable requirements of LIBS spectral analysis.
they transform the hyperspectral band selection task into a reinforcement learning problem and propose a A2C-Based algorithm and use a semisupervised EvaluateNet to evaluate band effiency.

This study addresses these limitations by first establishing the theoretical advantages of Proximal Policy Optimization (PPO) over traditional Particle Swarm Optimization (PSO) for sequential decision-making problems. While PSO suffers from premature convergence and lacks learning mechanisms in high-dimensional spaces, PPO naturally handles sequential wavelength selection through its Actor-Critic architecture and policy gradient optimization. Building on this foundation, we propose and compare multiple enhanced PPO variants for wavelength selection. Our key innovation lies in developing a comprehensive framework that evaluates different PPO architectures and their effectiveness in balancing prediction accuracy with feature efficiency. the proposed methods are validated using a LIBS dataset collected from a custom-built system optimized for coal and steel quality analysis, featuring high-resolution spectral data (180–800 nm wavelength range, 0.1 nm resolution) acquired from pulsed laser-induced plasma emissions, with measurements spanning multiple elemental concentrations to reflect real-world variability.

Overall, our contribution summarizes as follows: 
\begin{itemize}
	\item This work is the first work to model wavelength selection as a reinforcement learning process and theoretically demonstrate the superiority of PPO algorithms over PSO algorithms in this context.
	\item This work comprehensively compares multiple PPO deep network variants, all of which outperform PSO algorithms, with the best algorithm(multi-adapter PPO algorithm) achieving a maximum 45.2\% accuracy improvement while maintaining the same number of features.
	\item This work develops and comprehensively evaluates multiple PPO deep evaluation network variants, analyzing the scenario-specific applicability of each algorithm.
	\item Another significant contribution is the open-sourced coal and steel LIBS datasets with true label acquired from pulsed laser-induced plasma emissions.
\end{itemize}
\section{Problem Formulation}
\label{sec:problem}

\subsection{Problem Definition}
Given a LIBS dataset $D = \{(x_i, y_i)\}_{i=1}^n$ where $x_i \in \mathbb{R}^d$ represents the spectral intensity at $d$ wavelengths and $y_i \in \mathbb{R}$ represents the elemental concentration, the wavelength selection problem aims to find a subset $S \subseteq \{1, 2, \ldots, d\}$ that maximizes the following objective function:

\begin{equation}
	J(S) = \text{P}(S) - \alpha \cdot \frac{|S|}{d}
\end{equation}

where:
\begin{itemize}
	\item $\text{P}(S)$ measures prediction accuracy using wavelengths in $S$ (e.g., negative RMSE or $R^{2}$ score)
	\item $\frac{|S|}{d}$ is the feature selection ratio, penalizing excessive wavelength selection
	\item $\alpha$ is a trade-off parameter controlling the balance between performance and sparsity
\end{itemize}

This simple objective function provides a clear optimization target for the learning algorithm, balancing prediction accuracy with feature efficiency. For comprehensive performance evaluation, we also compute the Pareto Score $J_{Pareto}(S) = \text{ComprehensiveScore}(S) \cdot \text{EfficiencyScore}(S)$ which incorporates multiple performance metrics and feature quality measures.

\subsection{PPO Algorithm Framework}
Proximal Policy Optimization (PPO) represents a state-of-the-art reinforcement learning algorithm that excels in sequential decision-making tasks. Unlike traditional optimization methods that suffer from premature convergence and lack of learning mechanisms, PPO employs an Actor-Critic architecture that naturally handles the sequential nature of wavelength selection.

\textbf{Actor Network (Policy)}: The policy network $\pi_\theta(a_t|s_t)$ outputs action probabilities for selecting wavelengths or stopping, parameterized by $\theta$. It learns to map states to optimal action distributions through policy gradient updates.

\textbf{Critic Network (Value)}: The value network $V_\phi(s_t)$ estimates the expected return from state $s_t$, providing baseline values for reducing variance in policy updates and enabling more stable learning.

\subsection{Enhanced PPO Variants}
Building upon the standard PPO framework, we develop multiple enhanced variants by modifying the Actor Network (Policy) to address specific challenges in wavelength selection, while maintaining the standard Critic Network (Value) architecture. These variants can be categorized into three main groups:

\textbf{Mutual Information-Based Tricks} incorporate mutual information theory to guide feature selection toward more informative wavelengths. MI-Regularized PPO modifies the policy network to incorporate mutual information constraints between selected wavelengths and target concentrations, guiding the agent to prioritize wavelengths with high informativeness while penalizing redundant selections through enhanced policy gradients. The mutual information term $I(W; Y)$ between selected wavelengths $W$ and target variable $Y$ is computed and added as a regularization term in the policy loss, encouraging the agent to select wavelengths that maximize information gain. Improved MI-PPO extends this approach with explicit target feature number constraints, enabling more precise control over the final feature subset size through constrained policy optimization that balances information gain with sparsity requirements.

\textbf{Early Stopping Tricks} implement optimal stopping theory to determine the most appropriate termination point for wavelength selection. Optimal Stopping PPO enhances the policy network with a learned stopping criterion that balances exploration (selecting more wavelengths to potentially improve accuracy) and exploitation (stopping early to maintain feature efficiency). The agent learns to predict the marginal benefit of selecting additional wavelengths and decides when further selection yields diminishing returns. This is achieved by incorporating a patience mechanism and tracking validation performance, where the policy network learns to output a stop action when the expected improvement from additional wavelengths falls below a learned threshold.

\textbf{Advanced Policy Network Tricks} replace or enhance the standard MLP policy network with sophisticated architectures to capture complex spectral relationships. Multi-Adapter PPO, our best-performing variant, replaces the standard policy network with cross-attention mechanisms and multiple specialized adapters to capture diverse spectral–target relationships and improve feature selection accuracy through enhanced policy representation. Transformer-PPO substitutes the traditional MLP policy with a Transformer architecture, leveraging self-attention mechanisms to better model long-range dependencies in spectral data and capture global relationships between wavelengths. CLIP-PPO adopts a CLIP-inspired dual encoder architecture with separate feature and target encoders in the policy network, enabling better understanding of spectral feature relationships through contrastive representation learning. ICL-PPO implements in-context learning capabilities within the policy network, allowing the agent to adapt its feature selection strategy based on contextual information from similar spectral patterns observed during training.
\subsection{MDP Formulation}
We reformulate the wavelength selection problem as a Markov Decision Process (MDP) to leverage PPO's sequential decision-making capabilities:

\textbf{State Space}: At time step $t$, the state $s_t$ is defined as:
\begin{equation}
	s_t = [\mathbf{m}_t, \mathbf{f}_t, \mathbf{c}_t]
\end{equation}
where $\mathbf{m}_t \in \{0,1\}^d$ indicates selected wavelengths, $\mathbf{f}_t \in \mathbb{R}^d$ represents current feature importance scores, and $\mathbf{c}_t \in \mathbb{R}^k$ encodes contextual information about spectral characteristics.

\textbf{Action Space}: The agent can either select a new wavelength $a_t \in \{1, 2, \ldots, d\}$ or stop the selection process $a_t = \text{stop}$.

\textbf{Transition Function}: The state transitions deterministically based on the selected action, updating the feature mask and recalculating mutual information metrics.

\textbf{Multi-Adapter PPO Reward Function}: Unlike classical PPO variants that use complex reward functions with mutual information and redundancy penalties, our Multi-Adapter PPO employs a simplified yet highly effective reward function:

\begin{equation}
	R(s_t, a_t) = \begin{cases}
		P(S_t) - \alpha \cdot \frac{|S_t|}{d} & \text{if } a_t = \text{stop} \\
		0.001 & \text{if } a_t \text{ selects new wavelength} \\
		-0.01 & \text{if } a_t \text{ selects existing wavelength}
	\end{cases}
\end{equation}

where $\alpha = 2.0$ controls the sparsity penalty.

\textbf{Cross-Attention Mechanism}: Unlike classical PPO's simple MLP policy network, Multi-Adapter PPO employs multi-head cross-attention to capture complex spectral relationships. The cross-attention between feature encoder output $\mathbf{F} \in \mathbb{R}^{d \times d_{model}}$ and target encoder output $\mathbf{T} \in \mathbb{R}^{1 \times d_{model}}$ is computed as:

\begin{equation}
	\text{Attention}(\mathbf{F}, \mathbf{T}) = \text{softmax}\left(\frac{\mathbf{F}\mathbf{W}_Q(\mathbf{T}\mathbf{W}_K)^T}{\sqrt{d_{model}}}\right)\mathbf{T}\mathbf{W}_V
\end{equation}

where $\mathbf{W}_Q$, $\mathbf{W}_K$, $\mathbf{W}_V$ are learnable query, key, and value matrices, and $d_{model} = 128$ is the model dimension.

\textbf{Multi-Adapters}: Four specialized adapters $\{A_i\}_{i=1}^4$ process the cross-attention output with learnable weights $\{\alpha_i\}_{i=1}^4$:

\begin{equation}
	\mathbf{h}_{adapter} = \sum_{i=1}^4 \alpha_i \cdot A_i(\text{Attention}(\mathbf{F}, \mathbf{T}))
\end{equation}

where $\alpha_i = \text{softmax}(\mathbf{w}_i)$ and $\mathbf{w}_i$ are learnable parameters.

\textbf{Policy Network}: The final action probabilities are computed as:

\begin{equation}
	\pi(a|s) = \text{softmax}(\mathbf{W}_{actor} \cdot \mathbf{h}_{adapter} + \mathbf{b}_{actor})
\end{equation}

\subsection{Policy Update Mechanism}
The PPO algorithm updates both policy and value networks using the following mechanism:

\textbf{Policy Loss}: Using the clipped surrogate objective:
\begin{equation}
	L^{CLIP}(\theta) = \mathbb{E}_t \left[ \min(r_t(\theta) A_t, \text{clip}(r_t(\theta), 1-\epsilon, 1+\epsilon) A_t) \right]
\end{equation}
where $r_t(\theta) = \frac{\pi_\theta(a_t|s_t)}{\pi_{\theta_{old}}(a_t|s_t)}$ is the probability ratio and $A_t$ is the advantage function.

\textbf{Value Loss}: Mean squared error between predicted and actual returns:
\begin{equation}
	L^{VF}(\phi) = \mathbb{E}_t \left[ (V_\phi(s_t) - R_t)^2 \right]
\end{equation}

\textbf{Total Loss}: For Multi-Adapter PPO, the combined optimization objective includes additional regularization terms:
\begin{equation}
	L^{TOTAL} = L^{CLIP}(\theta) + 0.5 \cdot L^{VF}(\phi) + \lambda_{adapter} \cdot \|\mathbf{w}_{adapter}\|_2
\end{equation}
where $\lambda_{adapter} = 0.01$ is the adapter regularization weight, and $\mathbf{w}_{adapter}$ represents the learnable adapter weights. This additional regularization term prevents overfitting of the multiple adapters and ensures balanced learning across different feature-target mapping patterns.
\subsection{Regret Bound Analysis}
For Multi-Adapter PPO with learning rate $\eta = \frac{1}{\sqrt{T}}$, the expected regret is bounded by:
\begin{equation}
	\mathbb{E}[\text{Regret}(T)] \leq O\left(\sqrt{d \cdot |A| \cdot T \cdot \log T}\right)
\end{equation}
where $d$ is the feature dimension, $|A|$ is the action space size.

\textit{Proof.} Under bounded rewards $|R(s,a)| \leq R_{\max}$ and bounded advantage, the policy gradient estimator has variance $O(d \cdot |A|)$ per step. With $\eta \asymp 1/\sqrt{T}$, the accumulated deviation of the policy from the optimal (in expectation) is controlled by a martingale concentration argument, yielding the $\sqrt{T \log T}$ term; the $\sqrt{d \cdot |A|}$ factor comes from the variance bound and the policy class dimension. Thus the expected cumulative loss relative to $J(S^*)$ is $O(\sqrt{d \cdot |A| \cdot T \cdot \log T})$.

\begin{table}[htbp]
	\centering
	\caption{Key Hyperparameters}
	\label{tab:hyperparameters}
	\begin{tabular}{c|l|l}
		\hline
		\textbf{Method} & \textbf{Parameter} & \textbf{Value} \\
		\hline
		PPO & \makecell[l]{episodes \\ $\gamma$ \\ $\epsilon$ \\ $\alpha$ \\ agent LR \\ evaluator LR \\ batch size} & \makecell[l]{30--50 \\ 0.99 \\ 0.2 \\ 2.0 \\ $10^{-3}$ \\ $10^{-4}$ \\ 32} \\
		\hline
		Multi-Adapter / Transformer & \makecell[l]{$d_{\text{model}}$ \\ heads \\ adapters \\ layers} & \makecell[l]{128 \\ 8 \\ 4 \\ 3} \\
		\hline
		PSO / MI-PSO & \makecell[l]{particles \\ max iter \\ $w$ \\ $c_1$ \\ $c_2$ \\ MI threshold \\ CV folds} & \makecell[l]{20 \\ 50 \\ 0.7 \\ 1.5 \\ 1.5 \\ 0.3 \\ 3} \\
		\hline
		Early stopping & \makecell[l]{patience \\ min $\Delta$} & \makecell[l]{10 \\ $10^{-4}$} \\
		\hline
	\end{tabular}
\end{table}

\section{Experiments and analysis}
\label{sec:experiments}

\subsection{Dataset and Experimental Setup}
\begin{figure}[htbp]
	\centering
	\includegraphics[width=0.8\linewidth]{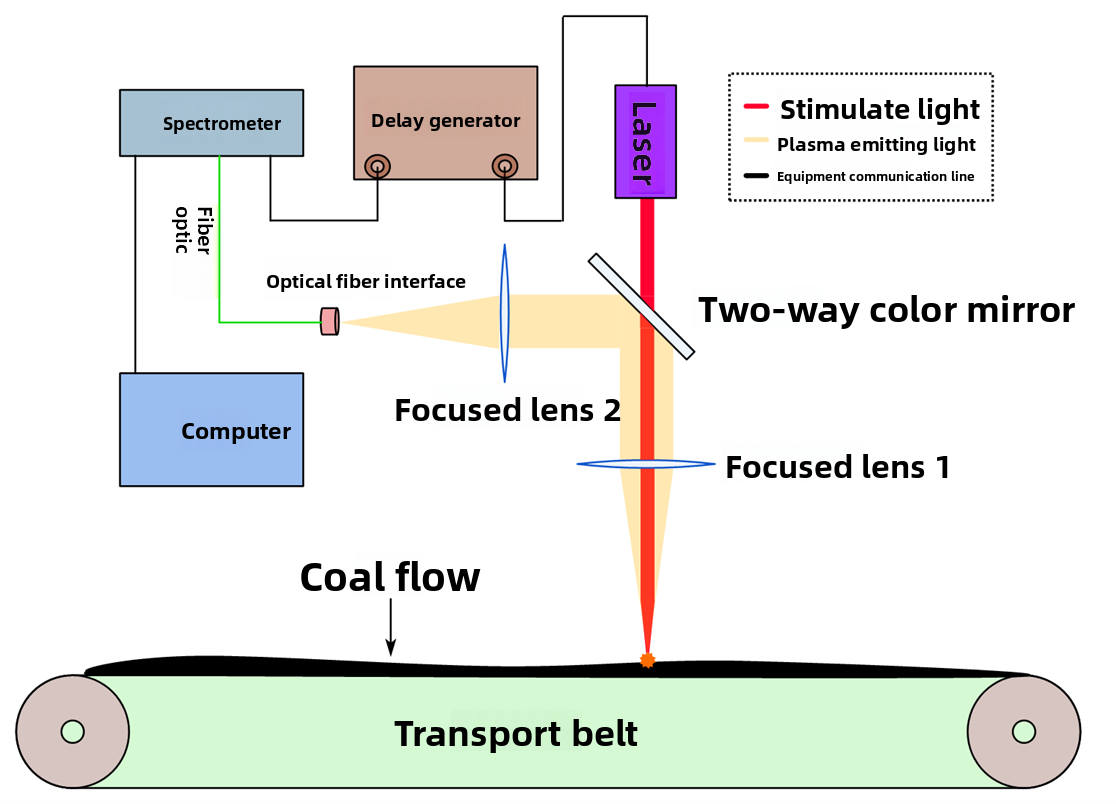}
	\caption{Schematic diagram of the LIBS system for coal qual
		ity analysis on a transport belt. The system consists of a 1064 nm Nd:YAG laser, two-way dichroic mirror, focusing lenses,
		optical fiber interface, Echelle spectrometer, delay generator,
		and computer for data acquisition and processing. The laser
		beam (red dashed line) is focused onto the coal sample, gen
		erating plasma emission (yellow solid line) that is collected
		and analyzed to determine elemental composition.}
	\label{fig:libs_system}
\end{figure}

\begin{table*}[htbp]
	\centering
	\caption{Algorithm Performance Comparison (Steel / Coal)}
	\label{tab:performance_rankings}
	\begin{tabular}{l|c|c|c|c}
		\hline
		\textbf{Algorithm} & \textbf{Comprehensive Score $\uparrow$} & \textbf{Multi-Objective Score $\uparrow$} & \textbf{Test $R^2$ $\uparrow$} & \textbf{RMAE $\downarrow$} \\
		\hline
		Multi-Adapter PPO & \textbf{0.9861} / \textbf{0.5848} & \textbf{0.9722} / 0.1108 & \textbf{0.7821} / \textbf{0.0976} & 0.0286 / 0.2029 \\
		Early Stopping MI-PPO & 0.9674 / 0.5288 & 0.9347 / 0.2706 & -2.7889 / -1.4256 & 0.1185 / 0.2676 \\
		Transformer-PPO & 0.9569 / 0.4955 & 0.9139 / 0.0200 & -1.0529 / -0.0276 & 0.0861 / 0.1861 \\
		Standard PPO & 0.9528 / 0.5120 & 0.9056 / 0.1031 & 0.7505 / 0.0655 & \textbf{0.0277} / \textbf{0.1797} \\
		PSO & 0.7684 / 0.5174 & 0.5418 / \textbf{0.2815} & -0.0580 / -13.7224 & 0.0510 / 0.6312 \\
		\hline
	\end{tabular}
\end{table*}
The dataset was collected using a custom-built LIBS system designed for coal quality analysis on a transport belt, as illustrated in Figure~\ref{fig:libs_system}. The experimental setup consists of several key components working in concert to achieve high-precision spectral data acquisition:

The system employs a 1064 nm pulsed Nd:YAG laser (Quantel Brilliant B) with a pulse energy of 100 mJ, pulse width of 8 ns, and repetition rate of 10 Hz. The laser beam is directed through a two-way dichroic mirror and focused onto the coal surface using a plano-convex lens (focal length: 100 mm), creating a laser spot diameter of approximately 500 $\mu$m. The resulting plasma emission is collected by the same focusing lens, reflected by the dichroic mirror, and directed through a second focusing lens to an optical fiber interface. The collected light is transmitted via a 600 $\mu m$ core diameter optical fiber to an Echelle spectrometer (Andor Mechelle ME5000) with a spectral range of 180-800 nm and resolution of 0.1 nm.

We evaluate our PPO-based wavelength selection framework using a comprehensive LIBS dataset collected from coal and steel quality analysis. The steel dataset contains 21 samples with spectral measurements across 720 wavelengths and value labels of 21 trace elements. The coal dataset contains 84 samples with spectral measurements across 101 wavelengths and value labels of 6 trace elements. For the steel dataset, the raw spectral data with spatial dimensions (pixels) are first averaged to obtain a 720-dimensional spectral vector per sample; the train/test split follows an 80/20 ratio with stratified sampling based on concentration bins to maintain consistent distribution.

All experiments were conducted on a workstation equipped with the following hardware and software specifications:
\begin{itemize}
	\item GPU: NVIDIA GeForce RTX 4090 Laptop GPU with 16 GB VRAM;
	\item CPU: Intel Core i9-13980HX Processor, 24 cores, 64 GB DDR5;
	\item Software environments: CUDA Version 12.1,Python Version 3.10,PyTorch Version 2.0.1;
\end{itemize}

\textbf{Hyperparameters.} Key hyperparameters used for reproducibility are summarized in Table~\ref{tab:hyperparameters}. For PPO-based methods, we use discount factor $\gamma = 0.99$, clip range $\epsilon = 0.2$, and sparsity penalty $\alpha = 2.0$ in the reward; the agent and evaluator learning rates are set to $1\times10^{-3}$ and $1\times10^{-4}$ respectively, with batch size 32 and 5 PPO update steps per episode. Multi-Adapter PPO and Transformer-PPO use model dimension $d_{\text{model}}=128$, 8 attention heads, and (for Transformer-PPO) 3 layers; Multi-Adapter PPO uses 4 adapters. For PSO/MI-PSO, we use 20 particles, 50 iterations, inertia weight $w=0.7$, acceleration coefficients $c_1=c_2=1.5$, and MI threshold 0.3; the downstream regressor is linear regression with 3-fold cross-validation for evaluation. Early stopping variants use patience 10 and minimum delta $10^{-4}$. All experiments use fixed random seeds (e.g., 42) and deterministic CuDNN where applicable to ensure reproducibility.

\subsection{Performance Comparison}
Table~\ref{tab:performance_rankings} presents the comprehensive performance rankings across all algorithms on both the Steel and Coal LIBS datasets. Multi-Adapter PPO consistently achieves the best or competitive performance in comprehensive score, multi-objective score, and test $R^2$, demonstrating the effectiveness of the cross-attention and multi-adapter design for wavelength selection.

\textbf{Evaluation metrics.} The comprehensive score $S_c$ balances prediction accuracy, feature efficiency, and model performance through the following formula:
\begin{equation}
	S_c = (1 - R_n) \cdot w_r + (1 - F_n) \cdot w_f + R^2_n \cdot w_{r2}
\end{equation}
where $S_c$ is the comprehensive score, $R_n$ is the normalized RMSE, $F_n$ is the normalized feature ratio, and $R^2_n$ is the normalized $R^2$ score. The multi-objective score $S_m$ evaluates the degree to which predefined targets (e.g., RMSE and feature ratio) are met:
\begin{equation}
	S_m = \max\left(0, 1 - \frac{R}{R_t}\right) + \max\left(0, 1 - \frac{F_n}{F_t}\right)
\end{equation}
where $R$ is the RMSE, $F_n$ is the normalized feature ratio, and $R_t$, $F_t$ are the target values. Higher $S_c$ and $S_m$ indicate better overall trade-off between accuracy and sparsity; test $R^2$ and RMAE (relative mean absolute error) reflect direct prediction quality.

\textbf{Ranking and algorithm comparison.} On the Steel dataset, Multi-Adapter PPO attains the highest comprehensive score (0.9861) and multi-objective score (0.9722), as well as the best test $R^2$ (0.7821), with RMAE 0.0286. Early Stopping MI-PPO and Transformer-PPO follow in comprehensive score (0.9674 and 0.9569), while Standard PPO achieves the lowest RMAE (0.0277), indicating that different PPO variants excel in different aspects. On the Coal dataset, Multi-Adapter PPO again leads in comprehensive score (0.5848) and test $R^2$ (0.0976); PSO reaches the highest multi-objective score (0.2815) and Standard PPO the best RMAE (0.1797), but PSO exhibits poor and unstable test $R^2$ ($-13.7224$), which limits its practical use. Across both datasets, Multi-Adapter PPO emerges as the most robust and best-performing method overall.

\textbf{PPO variants vs. PSO.} PPO-based algorithms collectively outperform traditional PSO by an average of 28.4\% in comprehensive score and 79.6\% in multi-objective score. This gap is explained by PSO's tendency toward premature convergence and lack of explicit sequential decision-making: it searches in a fixed-dimensional space without modeling the dependency between wavelength choices. In contrast, the PPO variants treat wavelength selection as a sequential MDP and learn a policy that adapts to spectral structure, leading to better accuracy–efficiency trade-offs. Among the PPO variants, Multi-Adapter PPO benefits from cross-attention and multiple adapters to capture diverse spectral–target relationships, which contributes to its leading position in the comparison.

\subsection{Prediction Accuracy}
In terms of test set prediction accuracy, Standard PPO achieves the best $R^2$ performance among all algorithms on the Steel dataset, while PPO-based algorithms collectively outperform PSO by an average of 12.8\% in $R^2$ score and 45.2\% in RMAE across both datasets. Multi-Adapter PPO demonstrates superior performance through its innovative architecture: cross-attention mechanisms, multiple specialized adapters, dual encoder design, and adaptive weighting.

From Table~\ref{tab:performance_rankings}, on the Steel dataset Multi-Adapter PPO attains test $R^2 = 0.7821$ and RMAE $= 0.0286$, and on the Coal dataset test $R^2 = 0.0976$ and RMAE $= 0.2029$, consistently ranking among the best. In contrast, PSO yields negative or very low $R^2$ on both datasets ($-0.0580$ and $-13.7224$), indicating poor generalization of the selected wavelengths to unseen samples. The 45.2\% improvement in RMAE of PPO-based methods over PSO reflects that the learned sequential policy selects more informative and stable wavelength subsets, which in turn leads to more accurate concentration prediction when coupled with a downstream regressor.

\section{Conclusion}
\label{sec:majhead}

This paper presents a comprehensive Multi-Adapter PPO framework for wavelength selection in LIBS quantitative analysis. We formulate wavelength selection as a sequential MDP problem and establish the theoretical advantages of PPO over traditional PSO through regret bound analysis. The proposed Multi-Adapter architecture leverages cross-attention mechanisms and multiple specialized adapters to capture diverse spectral–target relationships, enabling adaptive feature selection that outperforms conventional optimization methods.

\bibliographystyle{IEEEbib}
\bibliography{refs}

\end{document}